\documentclass[10pt,twocolumn,letterpaper]{article}

\usepackage[pagenumbers]{cvpr} 

\usepackage{graphicx}
\usepackage{subcaption} 
\usepackage{caption}
\usepackage{multirow}
\usepackage{makecell}
\usepackage{algorithm}
\usepackage{algorithmic}

\definecolor{cvprblue}{rgb}{0.21,0.49,0.74}
\usepackage[pagebackref,breaklinks,colorlinks,allcolors=cvprblue]{hyperref}

\def\paperID{*****} 
\def\confName{CVPR}
\def\confYear{2025}

\title{Noise Projection: Closing the Prompt–Agnostic Gap Behind Text-to-Image Misalignment in Diffusion Models}

\author{
Yunze Tong \quad Didi Zhu \quad Zijing Hu \quad Jinluan Yang \quad Ziyu Zhao \\
Zhejiang University \\
Hangzhou, Zhejiang, China \\
{\tt\small tyz01@zju.edu.cn}
}

\begin{document}
\maketitle

\begin{abstract}
In text-to-image generation, different initial noises induce distinct denoising paths with a pretrained Stable Diffusion (SD) model. While this pattern could output diverse images, some of them may fail to  align well with the prompt. Existing methods alleviate this issue either by altering the denoising dynamics or by drawing multiple noises and conducting post-selection. In this paper, we attribute the misalignment to a training–inference mismatch: during training, prompt-conditioned noises lie in a prompt-specific subset of the latent space, whereas at inference the noise is drawn from a prompt-agnostic Gaussian prior. To close this gap, we propose a noise projector that applies text-conditioned refinement to the initial noise before denoising. Conditioned on the prompt embedding, it maps the noise to a prompt-aware counterpart that better matches the distribution observed during SD training, without modifying the SD model. 
Our framework consists of these steps: we first sample some noises and obtain token-level feedback for their corresponding images from a vision–language model (VLM), then distill these signals into a reward model, and finally optimize the noise projector via a quasi-direct preference optimization. 
Our design has two benefits: (i) it requires no reference images or handcrafted priors, and (ii) it incurs small inference cost, replacing multi-sample selection with a single forward pass.Extensive experiments further show that our prompt-aware noise projection improves text-image alignment across diverse prompts.

\end{abstract}

\section{Introduction}\label{section: intro}

With the availability of large-scale data and powerful computing resources, diffusion models have emerged as highly effective generative frameworks. By learning to predict noise at varying levels, a diffusion model can start from pure Gaussian noise $x_t$ and iteratively denoise to reconstruct a clean image $x_0$. \citeauthor{score-based-method} further interpret this sampling process as a probability flow ordinary differential equation (ODE), where the only stochasticity arises from the initial noise. Consequently, any random noise sample can eventually be mapped to a clean image. To enable text-conditioned generation, Stable Diffusion (SD) \cite{stable-diffusion} incorporates text embeddings to guide the denoising trajectory, allowing outputting diverse images aligned with the input prompt from different random noises.

\begin{figure}[h]
    \centering
    \includegraphics[width=\linewidth]{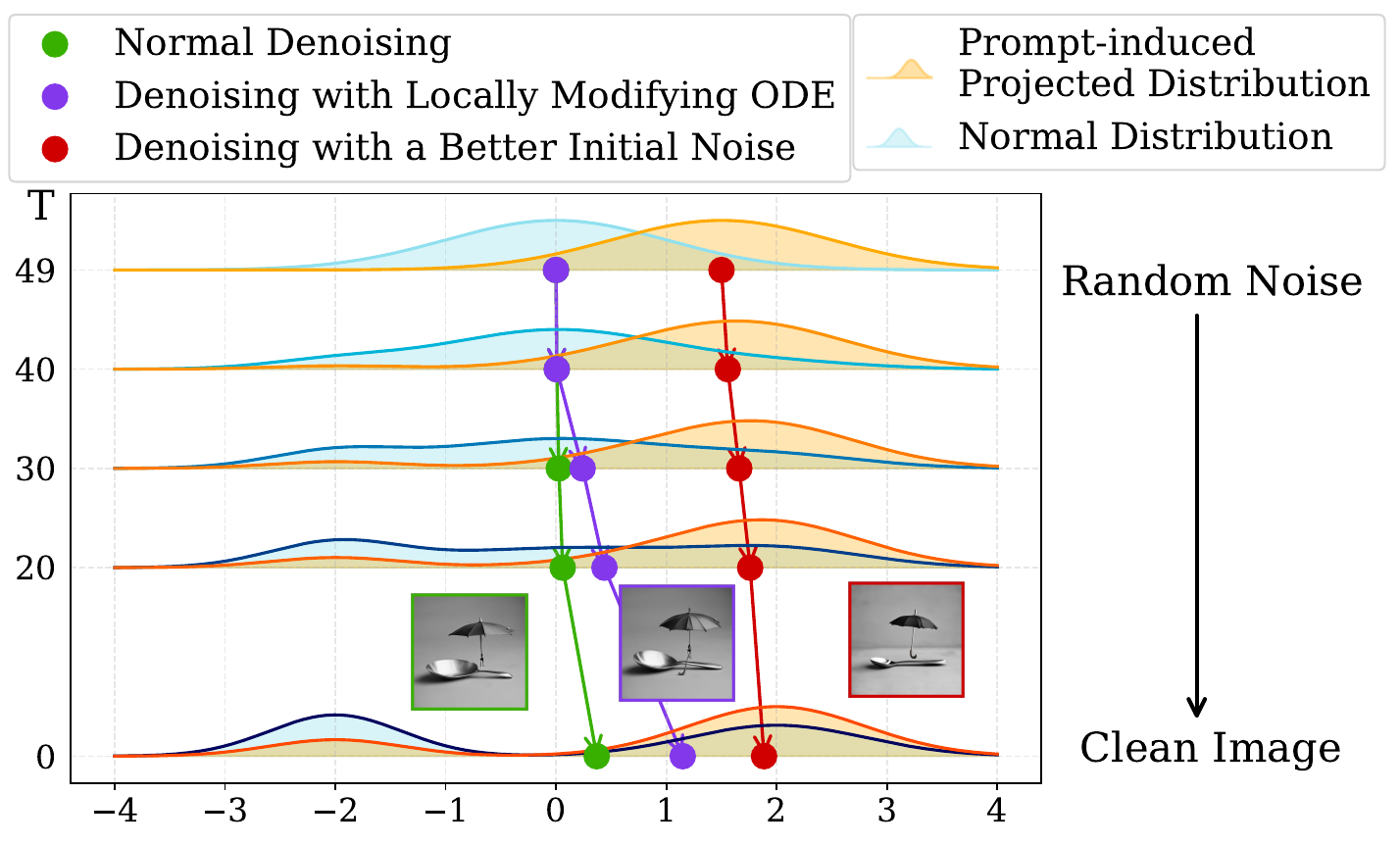}
    \caption{The comparison among several denoising patterns. Green path denotes normal denoising with a pretrained model, which has the risk of inducing text-image misalignment problem. Purple path reveals that optimization-based methods in essence modifies the ODE sampler locally. Red path denotes denoising from a better noise, which could be regarded as sampled from a prompt-conditioning distribution instead of normal Gaussian.}
    \label{fig: motivation}
\end{figure}

However, when sampling multiple images from the same prompt, different initial noises correspond to distinct ODE trajectories, leading to inconsistent text–image alignment, \textit{i.e.}, some samples faithfully match the prompt while others deviate. To address this, some optimization-based methods locally adjust the denoising path using reference images or human priors \cite{controlnet, ip-adapter, karras2024guidingdiffusionmodelbad, hu2025towards}, injecting auxiliary information to correct the ODE and reduce misalignment. These methods typically rely on external inputs and alter the denoising direction at every step. 
In contrast, sampling-based methods \cite{guo2024initno, miao2025noise, ma2025inference} pursue a global exploration strategy: they leave the denoising process unchanged but generate many candidates from diverse initial noise states through repeated sampling. The best-aligned images are then selected via human evaluation.
By covering a broader region of the distribution, these methods are more likely to yield images with stronger text–image alignment. However, this advantage comes at the expense of higher computational cost due to the large number of function evaluations required. 
We illustrate this comparison in Figure~\ref{fig: motivation}: optimization-based methods (purple) enhance alignment through stepwise modifications of the denoising process, typically by incorporating reference images during training or applying prior-guided interventions at inference; sampling-based methods instead select a suitable initial noise (red dot at $T=49$) from multiple random candidates.

In this paper, we aim to enhance text–image alignment by refining the original noise with a single projection rather than relying on multiple sampling. Concretely, we train a lightweight noise projector that takes the initial random noise and the text embedding as input, and produces a refined noise through a one-step propagation. Ideally, each initial noise—regardless of quality—can be mapped to a more suitable counterpart by the trained projector, which is then fed directly into the pretrained SD model. The key idea is to integrate text-conditioned information into the noise refinement process, thereby projecting the noise into a distribution that may deviate slightly from the Gaussian $\mathcal{N}(0,1)$ but aligns more closely with the given prompt.

Our motivation stems from the asymmetry between training and inference in SD. During training, prompts are mixed, and each noisy input is constructed by adding deterministic noise to clean images that exactly match the prompt. Thus, the noises available for each prompt form only a subset of all noisy inputs, and their implicit distribution may not follow $\mathcal{N}(0,1)$. Instead, it is the aggregate of noises across all prompts that conforms to the Gaussian distribution, illustrated as the blue area in Figure~\ref{fig: motivation}. At inference, however, generation is conditioned on a single deterministic prompt, while the initial noise is sampled from $\mathcal{N}(0,1)$ without prompt awareness. This mismatch can cause the sampled noise to deviate from the prompt-specific distribution observed during training, leading to poor alignment. To mitigate this, we introduce a noise projector that maps the initial noise toward the prompt-conditioned distribution, depicted as the yellow area in Figure~\ref{fig: motivation}.

To train the proposed noise projector, we leverage feedback from a pretrained Vision–Language Model (VLM). Given a set of prompts and seeds that determine the initial noises, we first generate images and obtain token-level scores from the VLM—one score per token per image—quantifying how strongly the image expresses the semantics of each prompt token (thus reflecting how well the initial noise realizes those semantics through the generation process). 
A reward model is then trained to approximate the VLM's scoring behavior. Finally, we adopt a quasi-direct preference optimization scheme to update the noise projector with the supervision from the reward model. 
The pipeline is fully automated, and optimization is confined to the reward model and the noise projector, whose parameter counts are far smaller than the SD backbone. 
Our design offers two key advantages: (i) training does not rely on human-provided reference images, nor does it impose constraints on the form of conditioning text, and (ii) inference incurs small overhead, since refining noise requires only a single forward pass through the noise projector without resorting to repeated sampling.

Our contributions are summarized as follows: (1) We analyze text–image misalignment from the perspective of initial noise, providing new insights into how it arises. (2) We propose a noise projector that converts a standard-Gaussian noise into a prompt-aware refined one, effectively steering it toward a prompt-conditioned distribution of the noise space and thereby improving alignment. (3) We develop a reinforcement-learning–based framework that uses VLM-proxied rewards to train the noise projector, eliminating the need for reference images during training and repeated sampling at inference. (4) Extensive experiments across diverse prompts validate the effectiveness of our method.

\section{Related Works}\label{section: related works}

\noindent {\bfseries Diffusion Models.} 
Diffusion models (DMs) have achieved remarkable success across diverse generative tasks \cite{SDXL, tabsyn, tong2025latent}. They typically adopt a UNet or Transformer backbone to estimate noise from corrupted inputs and progressively denoise toward a clean sample. \citeauthor{score-based-method} interpret this iterative process via a stochastic differential equation (SDE) and further derive a probability flow ordinary differential equation (ODE) that preserves the same marginal distribution. Latent diffusion models \cite{stable-diffusion} extend this framework to a compressed latent space, enabling efficient large-scale training. Moreover, advances in sampling strategies have accelerated inference \cite{DDIM, EDM} and facilitated conditional generation through guidance techniques \cite{CFG, karras2024guidingdiffusionmodelbad, kynkaanniemi2024applying}.

\noindent {\bfseries Improving Text-Image Alignment for DMs.}
For text-to-image generation, the primary goal is to ensure alignment between textual descriptions and synthesized images. However, without sufficient data scale or model capacity, DMs often fail under certain initial noise. To address this challenge, three major strategies have emerged: (1) scaling models or training datasets to improve coverage of the data distribution \cite{SDXL, flux2024}; (2) incorporating human priors \cite{fastcomposer, PAG, sundaram2024coconoattentioncontrastandcompleteinitial} or reference images \cite{ip-adapter, controlnet} to locally guide the denoising trajectory, achieved via fine-tuning or training-free integration; and (3) leveraging reward models or preference data to refine intermediate trajectories through direct preference optimization or reinforcement learning \cite{yang2025mix, hu2025d, zhou2025golden, eyring2024reno}. The first strategy expands the data space and modifies latent distributions, while the latter two primarily adjust the ODE dynamics implied by the pretrained model.

\section{Background}

\subsection{Latent Diffusion Models}

Latent Diffusion Models (LDMs) \cite{stable-diffusion} first compress images into latent representations using a pre-trained variational autoencoder (VAE). The diffusion process is then applied in the latent space for efficient modeling.
Denoising diffusion probabilistic models (DDPMs)~\cite{ddpm} define a forward process that gradually perturbs a clean data sample $\mathbf{x}_0$ into Gaussian noise through a sequence of conditional distributions. 
In closed form, the noisy sample at step $t$ is drawn from 
$q(\mathbf{x}_t \mid \mathbf{x}_0) = \mathcal{N}\!\big(\sqrt{\bar{\alpha}_t}\,\mathbf{x}_0,\;(1-\bar{\alpha}_t)\mathbf{I}\big)$,
where $\bar{\alpha}_t=\prod_{s=1}^{t}(1-\beta_s)$ and $\sigma^2(t)=1-\bar{\alpha}_t$.
Training reduces to learning a noise predictor $\epsilon_\theta(\mathbf{x}_t,t)$ that estimates $\epsilon$ in $\mathbf{x}_t = \sqrt{\bar{\alpha}_t} \mathbf{x}_0+ \sigma(t)\epsilon, \epsilon \sim \mathcal{N}(0,I)$. 
\citeauthor{score-based-method} presented a continuous-time formulation for this variance-preserving diffusion, which corresponds to the stochastic differential equation (SDE)
\begin{equation}\label{eq: SDE for DDPM forward}
    d\mathbf{x}_t = -\tfrac{1}{2}\beta(t)\,\mathbf{x}_t\,dt + \sqrt{\beta(t)}\,d\mathbf{w}_t.
\end{equation}
The reverse process is interpreted as iterative denoising, where the model $\epsilon_\theta(\mathbf{x}_t,t)$ reconstructs $\mathbf{x}_{0}$ from $\mathbf{x}_t$. 
The reverse-time dynamics of Eq.~\ref{eq: SDE for DDPM forward} yield the generative process. An equivalent deterministic formulation is given by the 
probability-flow ordinary differential equation (ODE):
\begin{equation}\label{eq: ODE reverse}
    \tfrac{d\mathbf{x}_t}{dt} = -\tfrac{1}{2}\beta(t)\,\mathbf{x}_t + \tfrac{1}{2}\beta(t)\,\tfrac{1}{\sigma(t)}\,\epsilon_\theta(\mathbf{x}_t,t),
\end{equation}
which shares the same marginals as the SDE.  
In practice, modern pretrained models retain the DDPM-style forward training objective, while inference relies on integrating the ODE sampler with efficient solvers (\textit{e.g.}, DDIM~\cite{DDIM}, DPM-Solver~\cite{lu2022dpm}, or Euler ancestral methods~\cite{EDM}). By modifying the ODE dynamics, some works enable diverse sampling behaviors tailored to their specific tasks.

\subsection{Achieving Text-Image Alignment}

With Eq.~\ref{eq: ODE reverse}, diverse images can be generated. However, such unconditional sampling lacks text guidance, so the generated results may not reflect the desired semantics. To obtain samples with specific labels, classifier-free guidance (CFG) \cite{CFG} is widely adopted for efficient text conditioning. In this setting, a text embedding $\mathbf{c}$ is provided as input, yielding the noise estimator $\epsilon_\theta(\mathbf{x}_t, t, \mathbf{c})$. 
A single model thus supports predicting the denoising directions for both conditional and unconditional cases through $\epsilon_\theta(\mathbf{x}_t, t, \mathbf{c})$ and $\epsilon_\theta(\mathbf{x}_t, t, \varnothing)$, respectively. 
During denoising, the effective output is defined as $\tilde{\epsilon}_\theta(\mathbf{x}_t, t, \mathbf{c}) = (1+w)\epsilon_\theta(\mathbf{x}_t, t, \mathbf{c}) - w\epsilon_\theta(\mathbf{x}_t, t, \varnothing)$,
which steers generation toward the desired prompt.

\subsection{Motivation of Projecting Noise}\label{subsection: motivation of projecting noise}

In text-to-image (T2I) generation, achieving strong alignment between text and visual output is critical. However, for challenging prompts or rare visual concepts, standard CFG often fails to provide sufficient guidance, and pretrained models may struggle to produce well-aligned samples. To mitigate this issue without incurring heavy retraining costs, prior works integrate additional information from reference images or human-defined priors. Such techniques, whether through fine-tuning or sampling interventions, can be interpreted as modifications to the ODE sampler in Eq.~\ref{eq: ODE reverse}, where auxiliary information is injected during denoising. While effective, these optimization-based methods typically require extra inputs and careful hyperparameter tuning. 

An alternative direction enhances alignment without external priors by sampling multiple candidates. These approaches generate outputs from diverse initial noises or by repeating sampling during inference, followed by evaluation to select the best candidate. Unlike ODE-modification methods, they leave the pretrained denoising dynamics unchanged and instead enlarge the search space of initial noise. The observed improvement in text-image alignment arises from post-selection: well-aligned samples correspond to a subset of initial noises that form an implicit posterior distribution conditioned on the prompt. 
To better illustrate this phenomenon, consider two sampled noises, $\epsilon_0$ and $\epsilon_1$. For each, we generate images under two different text prompts $\mathbf{c_0}$ and $\mathbf{c_1}$, denoted as $x_{\epsilon_0, \mathbf{c_0}}, x_{\epsilon_0, \mathbf{c_1}}, x_{\epsilon_1, \mathbf{c_0}}, x_{\epsilon_1, \mathbf{c_1}}$. It is possible that $x_{\epsilon_0, \mathbf{c_0}}$ better aligns with $\mathbf{c_0}$ than $x_{\epsilon_1, \mathbf{c_0}}$, while the reverse holds for $\mathbf{c_1}$. This phenomenon can be understood via Eq.~\ref{eq: ODE reverse}: during sampling, text conditions $\mathbf{c}$ can pair arbitrarily with noises $\epsilon$, whereas during training, noisy inputs are constructed by adding sample-specific deterministic noise at varying scales. As a result: (1) the effective noise space during training is narrower than that of fully random Gaussian noise; and (2) each text condition is only observed with the noise realizations derived from its paired training images. Consequently, at inference time, well-aligned generations correspond only to certain regions of the noise space—a subset of the Gaussian prior—thus defining a prompt-dependent noise distribution. In other words, producing semantically faithful images implicitly requires sampling from a unique, condition-specific distribution.

We use Figure~\ref{fig: motivation} to illustrate the difference between optimization-based and sampling-based methods. Without introducing additional priors or altering the data distribution, optimization-based methods can be viewed as locally modifying the ODE, thereby altering the denoising trajectory, as indicated by the purple line. Since denoising proceeds step by step, errors made in early stages propagate, so these methods require the trajectory to remain accurate throughout; otherwise, a local deviation may cause complete failure. In contrast, sampling-based methods aim to select better initial noises that align with the given condition. Statistically, selected noises are more likely to lie within favorable regions of the distribution, making the subsequent denoising more likely to yield aligned outputs. However, this comes at the cost of multiple function evaluations during inference. Motivated by this trade-off, we ask: can we simplify sampling-based methods into a faster optimization-based approach that achieves accurate alignment with lower inference cost?  

To this end, we propose a \textbf{noise projector} that utilizes text guidance before denoising begins. The projector is designed to map any randomly sampled noise to a refined noise. Once trained, it directly improves text-image alignment while avoiding repeated sampling during inference. Moreover, the projector operates independently of the standard SD pipeline, requiring no modification of pretrained model parameters.

\section{Method}\label{section: method}

\subsection{Model Architecture}\label{subsection: model arch}

\begin{figure}[h]
    \centering
    \includegraphics[width=\linewidth]{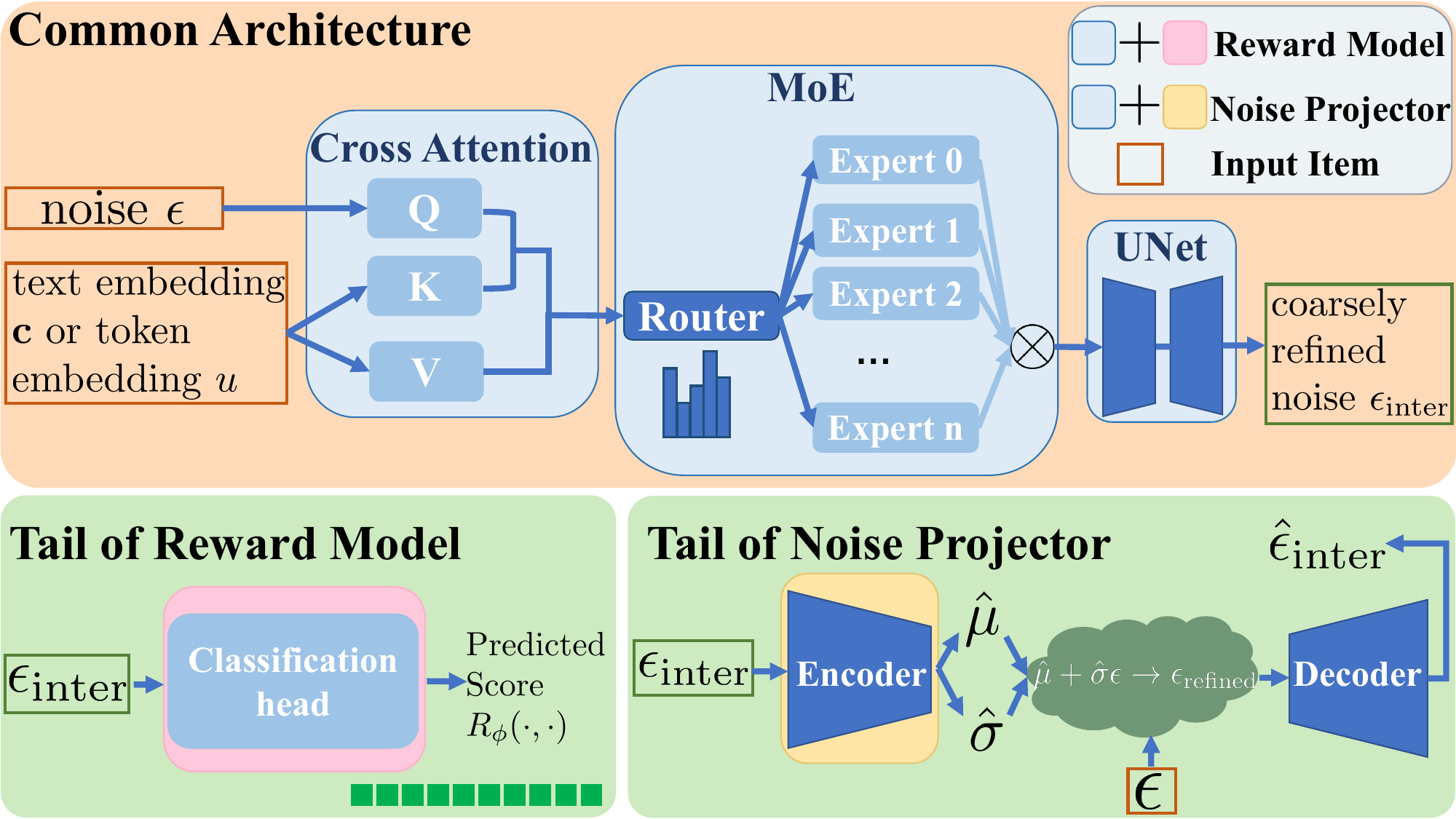}
    \caption{
    The architecture of our model. Both the noise projector and the reward model share the same backbone with cross-attention, MoE, and UNet components. At the output stage, the reward model attaches a classification head, while the noise projector integrates the encoder of a VAE.
    }
    \label{fig: model architecture}
\end{figure}

Our method involves training two models: a \textit{noise projector}, which maps the original noise to a refined one with improved text-image alignment, and a \textit{reward model}, which provides supervision signals to train the projector. Both take a noise sample and a text embedding as input, sharing the same backbone architecture in the early layers but differing in their output heads to match task-specific objectives. The overall design is illustrated in Figure~\ref{fig: model architecture}. 
The backbone begins with a cross-attention module to couple noise and text, producing \textit{mixed latents} that encode both semantic and stochastic information. The latents are then processed by a Mixture of Experts (MoE), where the router selectively activates experts to disentangle different semantic components. The MoE output represents a projected latent that integrates text-conditioned semantics. Finally, a UNet module reconstructs the noise layout from the projected latents.

Beyond these shared modules, the noise projector and reward model incorporate task-specific output heads. Their detailed designs are described below.

\subsubsection{Noise Projector}

The noise projector refines an input noise sample conditioned on text. The text input is the embedding of the full prompt, identical to the conditional input used in Stable Diffusion. This embedding conveys the complete semantic context, while the mapping from text tokens to pixel-level noise primarily occurs in the MoE. Unlike a standard UNet output, we append an auto-encoder that predicts both $\mu$ and $\sigma$, from which the final refined noise is sampled via reparameterization with $\epsilon_{\text{init}}$. This design prevents the refined noise from drifting too far from the $\mathcal{N}(0,1)$ initialization, which could otherwise lead to invalid images during early training. Details are provided in Section~\ref{subsubsection: pretrain noise projector}.

\subsubsection{Reward Model}

Unlike the noise projector, the reward model conditions on the embedding of a single token rather than the entire sentence, enabling token-level feedback and reducing reward sparsity which will be discussed in Section~\ref{subsection: train reward model}. Built on the shared backbone, it incorporates an extra MLP and a classification head, producing a normalized probability distribution aligned with the discrete scoring format.

\subsection{Training a Reward Model}\label{subsection: train reward model}
To guide the noise projector toward generating noise distributions that better align with the conditional prompt, obtaining reliable supervision signals is crucial. We use a large Vision Language Model (VLM) as the source of rewards due to its strong semantic understanding. However, the large parameter size of VLMs makes them impractical to involve in training. Therefore, we train a smaller reward model that approximates the predictive behavior of the VLM. This reward model acts as a proxy, enabling efficient training of the noise projector without requiring human evaluation, and can be executed fully automatically.

\subsubsection{Preparing Data}\label{subsubsection: prepare data for reward model}

We begin by generating a set of noises from predefined seeds and running the Stable Diffusion (SD) pipeline to obtain their corresponding images conditioned on prompts. Each image, together with one semantic token from its prompt, is then fed into the VLM. The VLM assigns a discrete score from 0 to 9, reflecting how well the image (and thus the noise) represents the given token. By traversing all semantically meaningful tokens in the prompt, we obtain a batch of token-level scores for each noise sample. This yields training pairs of the form $\{\epsilon_i, u_j, s_{ij}\}_{i,j}$, where $\epsilon_i$ denotes the $i$-th noise, $u_j$ the embedding of the $j$-th token, and $s_{ij}$ the score measuring how well $\epsilon_i$ aligns with $u_j$.

\subsubsection{Aligning Reward Model with VLM}\label{subsubsection: align reward model}

With these token-level pairs, we train a reward model $R_{\phi}$ to approximate the VLM’s judgments. Since the scores are discrete values between 0 and 9, this task can be formulated as multi-class classification. Specifically, the classification head of $R_{\phi}$ outputs a 10-dimensional vector, \textit{i.e.}, $R_{\phi}(\epsilon, u) \in \mathbb{R}^{10}$. We then optimize $R_{\phi}$ using cross-entropy loss over the collected pairs:
\begin{equation}\label{eq: loss for reward model}
    \mathcal{L}_{\text{RM}} = \sum_{i,j} \ell_{\text{CE}}(R_{\phi}(\epsilon_i, u_j), s_{ij}).
\end{equation}
After convergence, $R_{\phi}$ provides an efficient and faithful proxy to the VLM, offering dense token-level supervision for training the noise projector.

\subsection{Training the Noise Projector}

With a well-trained reward model $R_{\phi}$, we proceed to train the noise projector $P_{\theta}$.

\subsubsection{Pretraining}\label{subsubsection: pretrain noise projector}

\begin{figure*}[h]
    \centering
    \includegraphics[width=.9\linewidth]{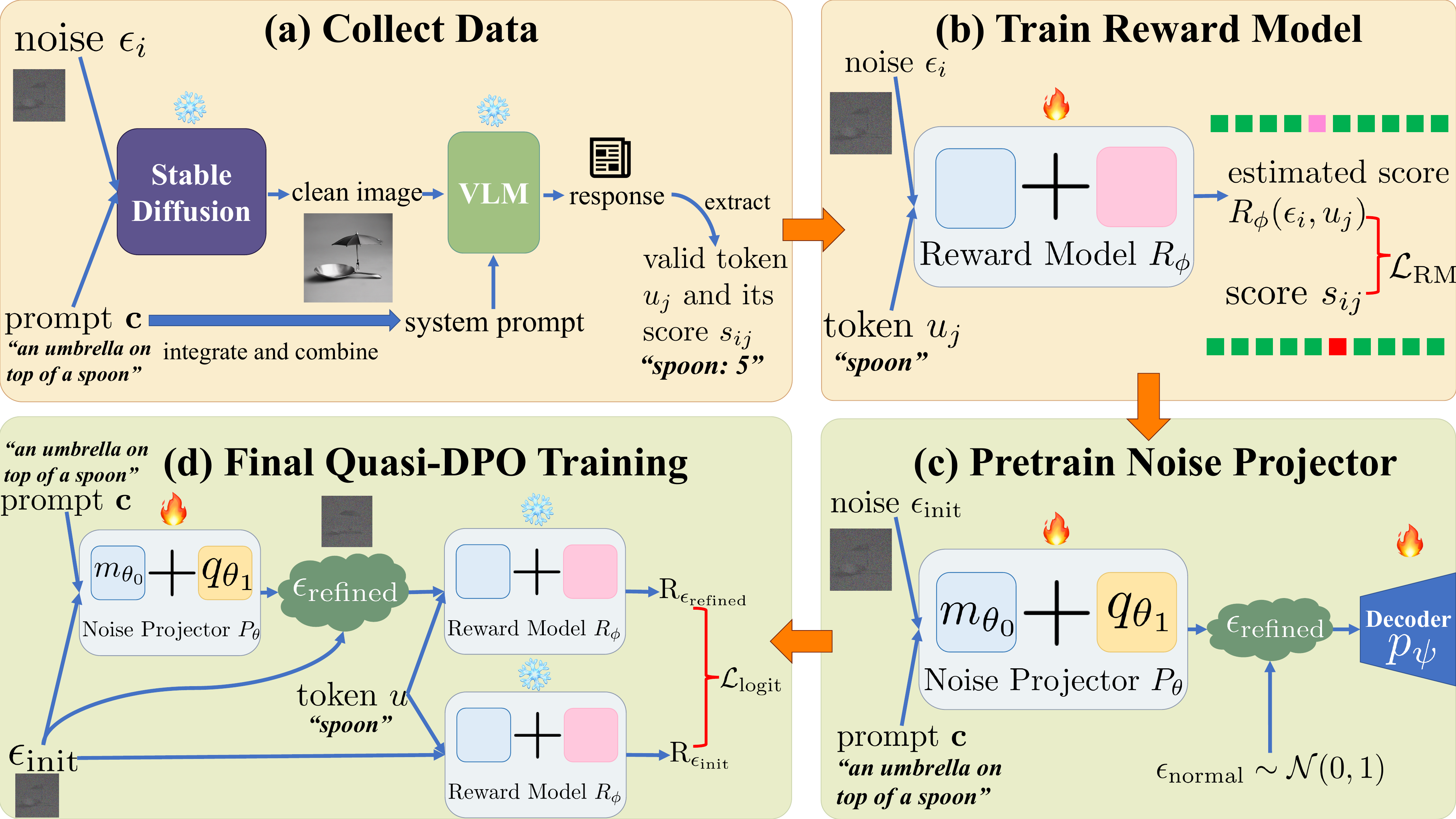}
    \caption{
    The overall framework of our method, which consists of four stages: (a) data preparation for training the reward model (Section~\ref{subsubsection: prepare data for reward model}), (b) reward model training with the collected data (Section~\ref{subsubsection: align reward model}), (c) pretraining the noise projector (Section~\ref{subsubsection: pretrain noise projector}), and (d) final quasi-DPO training (Section~\ref{subsubsection: final training}).
    }
\end{figure*}

As discussed in Section~\ref{subsection: motivation of projecting noise}, for a given conditional prompt, the set of effective noises that lead to well-aligned images forms a distribution that is distinct from, yet close to, the standard Gaussian $\mathcal{N}(0,1)$. The goal of $P_{\theta}$ is to map arbitrary input noise into this refined distribution.  

However, a challenge arises in the early stage of training: a randomly initialized $P_{\theta}$ may project noise far away from $\mathcal{N}(0,1)$. When the deviation is too large, the reward model $R_{\phi}$ cannot provide effective optimization signals because such highly deviated noise lies outside its training distribution. In this case, the optimization becomes unstable, and the projected noise may even fail to produce valid images after denoising, as the pretrained SD model requires inputs close to $\mathcal{N}(0,1)$. To resolve this, we introduce a pretraining stage to stabilize $P_{\theta}$ before reinforcement learning.

As described in Section~\ref{subsection: model arch}, the noise projector can be decomposed as $P_{\theta} = m_{\theta_0} \cdot q_{\theta_1}$. Here, $m_{\theta_0}$ includes the cross-attention, MoE, and UNet modules, while $q_{\theta_1}$ is the encoder of a variational autoencoder (VAE) with decoder $p_{\psi}$. Given a text embedding $\mathbf{c}$ and an initial noise $\epsilon_{\text{init}}$, $P_{\theta}$ outputs $\hat{\mu}$ and $\hat{\sigma}$ of the same shape as $\epsilon_{\text{init}}$. The refined noise is then sampled via the reparameterization trick: $\epsilon_{\text{refined}} = \hat{\mu} + \hat{\sigma} \odot \epsilon_{\text{normal}},  \epsilon_{\text{normal}} \sim \mathcal{N}(0,1)$.

To prevent $\epsilon_{\text{refined}}$ from drifting too far from $\mathcal{N}(0,1)$, we regularize the posterior defined by $(\hat{\mu}, \hat{\sigma})$ with a KL loss:
\begin{equation}\label{eq: KL loss for warm up}
    \mathcal{L}_{\text{constraint}} = \tfrac{\lambda}{2}\big(\hat{\mu}^2 + \hat{\sigma}^2 - 2\log\hat{\sigma} - 1\big),
\end{equation}
which encourages $\hat{\mu}\!\approx\!0$ and $\hat{\sigma}\!\approx\!1$. Consequently, since $\epsilon_{\text{normal}} \sim \mathcal{N}(0,1)$, the refined noise $\epsilon_{\text{refined}}$ remains close to the standard Gaussian distribution.

To ensure that the VAE captures the information from $m_{\theta_0}(\epsilon_{\text{init}}, \mathbf{c})$, we also apply a reconstruction loss:
\begin{equation}\label{eq: MSE loss for warm up}
    \mathcal{L}_{\text{reconstruction}} = \ell_{\text{MSE}}\!\big(m_{\theta_0}(\epsilon_{\text{init}}, \mathbf{c}),\; p_{\psi}(\epsilon_{\text{refined}})\big).
\end{equation}
This ensures that the VAE reconstructs the intermediate refined noise $m_{\theta_0}(\epsilon_{\text{init}}, \mathbf{c})$, with information consistently propagated through both $q_{\theta_1}$ and $p_{\psi}$. 

Finally, the pretraining stage jointly optimizes the noise projector and the decoder of VAE with:
\begin{equation}\label{eq: total loss for warm up}
    \mathcal{L}_{\text{warmup}} = \mathcal{L}_{\text{constraint}} + \mathcal{L}_{\text{reconstruction}}.
\end{equation}
Eq.~\ref{eq: total loss for warm up} ensures that projected noise remains close to the Gaussian prior while retaining semantic information, thereby stabilizing subsequent RL-based training. After pretraining, we discard the decoder $p_{\psi}$ and carry $P_{\theta}$ into the next stage.

\subsubsection{Final Training}\label{subsubsection: final training}

With the reward model $R_\phi$ trained, we now optimize the noise projector $P_{\theta}$. The training input includes noises, each determined by a single seed within a fixed range, along with conditioning prompts. 
For each pair $\{\epsilon_{\text{init}}, \mathbf{c}\}$, we obtain a projected noise $\epsilon_{\text{refined}}$ via reparameterization:
$\epsilon_{\text{refined}} = \hat{\mu} + \hat{\sigma} \odot \epsilon_{\text{init}}$. Using $\epsilon_{\text{init}}$ directly in the reparameterization offers two advantages: (1) it already follows $\mathcal{N}(0,1)$, avoiding redundant resampling; and (2) it preserves the structural characteristics of the original noise. If $\epsilon_{\text{init}}$ is already well aligned, then ideally $\hat{\mu}\!\to\!0$ and $\hat{\sigma}\!\to\!1$, yielding $\epsilon_{\text{refined}} \approx \epsilon_{\text{init}}$ and preventing unnecessary modifications.

For each noise pair $(\epsilon_{\text{init}}, \epsilon_{\text{refined}})$, the reward model outputs $R_\phi(\epsilon_{\text{init}}, u), R_\phi(\epsilon_{\text{refined}}, u) \in\mathbb{R}^{10}$, representing normalized distributions over discrete scores, where the $0$-th entry indicates the worst alignment and the $9$-th entry the best. We convert this to a scalar reward by multiplying with $v=(0,1,\dots,9)^\top \in \mathbb{R}^{10}$: $\mathrm{R}_{\epsilon} = R_\phi(\epsilon, u)v$.
We then adopt a quasi-direct preference optimization (DPO) objective:
\begin{equation}\label{eq: unweighted DPO loss}
    \mathcal{L}_{\text{unweighted}} = \log\!\big(1+ \exp(-(\mathrm{R}_{\epsilon_{\text{refined}}}-\mathrm{R}_{\epsilon_{\text{init}}}))\big).
\end{equation}
This contrastive formulation encourages $P_{\theta}$ to increase rewards relative to the refined noise. Although $\mathrm{R}_{\epsilon_{\text{init}}}$ does not depend on $P_{\theta}$, including it in the loss reweights samples according to their initial quality, similar to the role of the reference model in standard DPO. In this way, variations in the alignment quality of original noises are used to scale the incremental reward contributed by $P_\theta$.

To further account for reward magnitude, we deploy an extra reweighting scheme. Let $r(\epsilon)$ denote the discrete score index assigned to $\epsilon$ (i.e., the argmax of $R_\phi(\epsilon)$). Intuitively, samples with low scores (e.g., $r=0$) require stronger optimization than those already judged as well-aligned ($r=9$). We therefore define a weight vector $w \in \mathbb{R}^{10}$:
\[
w[i] = 1 + w_{\text{max}} - {w_{\text{max}}}^{\tfrac{i}{9}}, \quad i=0,\dots,9,
\]
where $w_{\text{max}}$ is a hyperparameter (set to 5). The weighted objective becomes
\begin{equation}
    \mathcal{L}_{\text{logit}} = \!\!\!\!\sum_{\{\epsilon_{\text{init}}, \mathbf{c}\}}\!\!\!\! w[r_{\epsilon_{\text{refined}}}] \cdot \log\!\big(1+ \exp(-\beta(\mathrm{R}_{\epsilon_{\text{refined}}}-\mathrm{R}_{\epsilon_{\text{init}}}))\big).
\end{equation}
We reuse Eq.~\ref{eq: KL loss for warm up} to prevent severe deviation of the noise distribution. The final training objective becomes:
\begin{equation}\label{eq: final loss for RL}
    \mathcal{L}_{\text{final}} = \mathcal{L}_{\text{logit}} + \tau \mathcal{L}_{\text{constraint}},
\end{equation}
where $\tau$ balances noise refinement with maintaining proximity to the standard Gaussian distribution.
Unlike standard DPO, our optimization relies on a reward model, and gradients flow only through the refined branch. Crucially, since $R_\phi$ is trained with token-level supervision from a VLM (Section~\ref{subsubsection: prepare data for reward model}), our rewards are significantly denser than sentence-level signals, providing more effective guidance for training the noise projector.

\begin{figure*}[h]
    \centering
    \includegraphics[width=.9\linewidth]{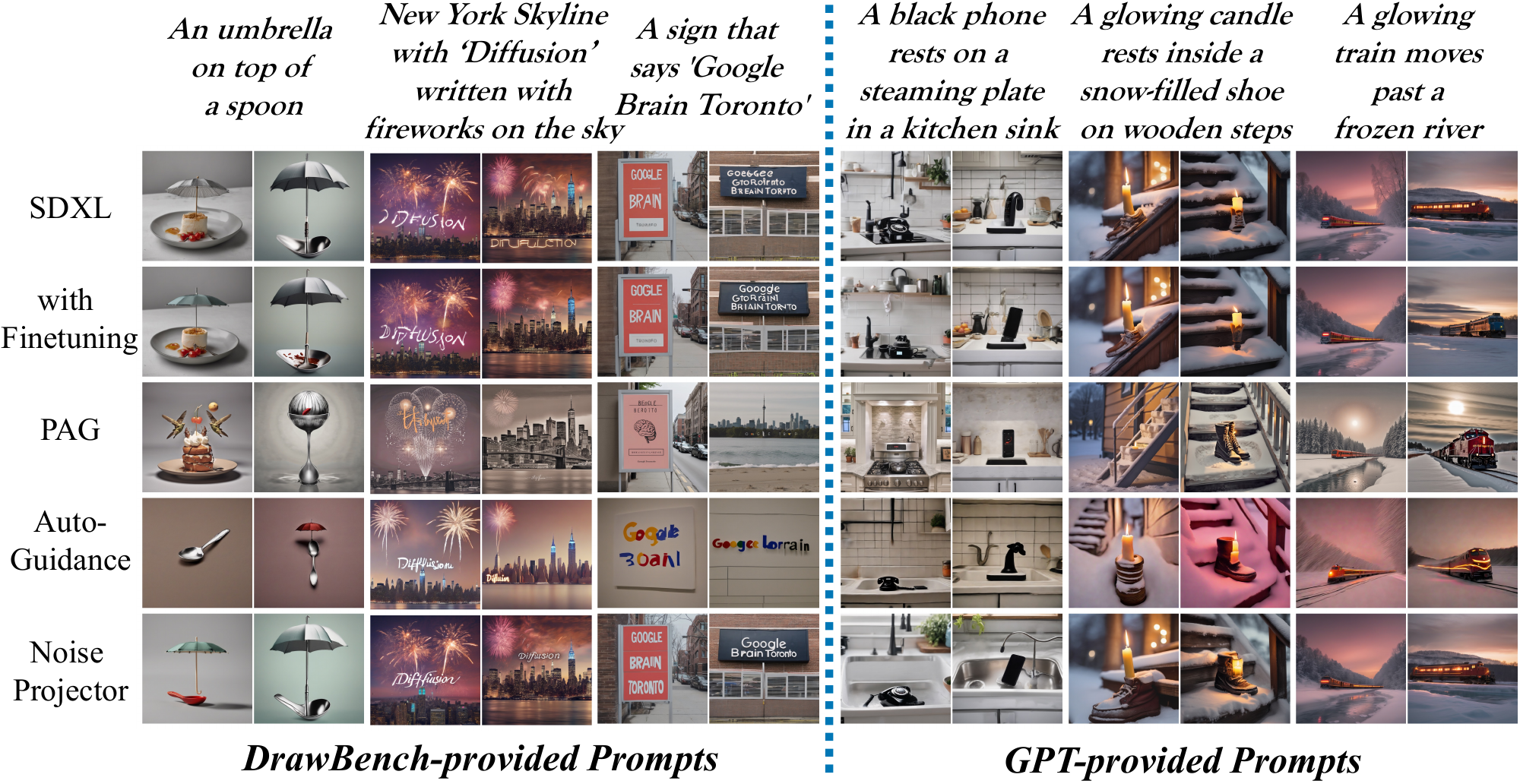}
    \vspace{-10pt}
    \caption{
    Qualitative comparison of generated images. For complex prompts that often induce text–image misalignment, our noise projector effectively refines the noise to yield well-aligned generations. In cases where the original SDXL already produces satisfactory results (\textit{i.e.}, the rightmost column), our noise projector leaves the structure largely unchanged, preserving the same well-aligned layout as the original.
    }
\end{figure*}

\section{Experiment}

\subsection{Experimental Settings}

\noindent \textbf{Evaluation Metrics.}
Our objective is to assess the alignment between generated images and their conditioning prompts. We adopt three evaluation protocols:

\begin{itemize}
    \item \textit{QwenScore}: 
    Conventional reward models take an image-prompt pair as input and directly output a scalar score to indicate the alignment, but such judgments are limited by the scope of their training data. In contrast, large vision–language models (VLMs) are trained on massive web-scale corpora and further enhanced by instruction tuning, which equips them with stronger semantic understanding and the ability to follow fine-grained evaluation instructions. We therefore query the VLM for discrete scores between 0 and 99 to measure image–prompt alignment. Specifically, we adopt the instruction-tuned Qwen2.5-VL-7B \cite{bai2025qwen25vltechnicalreport} as the evaluator, with the full instruction prompt detailed in the Appendix. 

    \item \textit{BERTScore}: 
    BERTScore \cite{DDPO} is a text-based metric for evaluating text-image alignment. Given an image, a textual description is first generated using a VLM, and the semantic similarity between the description and the original prompt is then computed. The similarity is quantified using the recall metric of BERT \cite{BERT}. For implementation, we employ instruction-tuned Qwen2.5-VL-7B to generate captions and DeBERTa-XLarge \cite{he2021deberta} to obtain text embeddings for similarity calculation.

    \item \textit{ImageReward}: 
    ImageReward \cite{imagereward} is a reward model trained on human-preference data. It directly takes images as input and produces scalar scores indicative of human judgments of quality and preference.
\end{itemize}

\noindent \textbf{Dataset.} Our paradigm constructs a background dataset by fixing a range of random seeds, each mapped to a Gaussian noise sample, and generating the corresponding images with a pretrained Stable Diffusion (SD) model. This process requires no external human-provided images and can be applied to any problematic conditioning prompt. Once the target prompts are specified, the dataset is constructed automatically.  
The prompts in our experiments are drawn from two sources and evaluated separately: (1) \textit{DrawBench} \cite{drawbench}, which offers challenging cases for assessing text–image alignment; and (2) GPT-4o \cite{openai2024gpt4ocard}, which we explicitly query to provide diverse prompts that are likely to induce misalignment, such as object omission, spatial confusion, and underrepresented textual content.  
To assess the effectiveness of the noise projector, we consider two evaluation settings: (i) \textit{single-prompt}, where the projector is trained for one specific prompt, and (ii) \textit{multi-prompt}, where a single projector is trained to handle multiple prompts simultaneously.

\noindent \textbf{Implementation Details.} 
We adopt SDXL \cite{SDXL} as the base model to generate images at a resolution of 1024 $\times$ 1024, and conduct all experiments on NVIDIA A100 GPUs. The framework involves training two components: a reward model and the final noise projector, with architectural details provided in the Appendix. For the reward model, training inputs are constructed from noises generated by seeds in the range $[0,300)$. For the noise projector, the training set includes noises from $[0,50)$ in the single-prompt setting and $[0,100)$ in the multi-prompt setting. The hyperparameters $w_\text{max}$ and $\tau$ are set to 5 and 200, respectively, and gradient norms are clipped to stabilize training.   

Since both our method and several baselines require training, we evaluate performance on both \emph{seen} and \emph{unseen} noises. Results on seen noises reflect the ability to fit the training distribution \cite{yang2024discovering, yang2025unifying}, while results on unseen noises indicate generalization to a wider range \cite{zhu2025remedy, yang2024mitigating, yang2025leveraging}. Among the two, performance on unseen data serves as the primary evaluation protocol, as it demonstrates the projector’s capability to handle arbitrary noises. Seen-data performance is measured using the same (or a subset of) noise samples employed in projector training, whereas unseen-data performance is measured on a disjoint range of random seeds not accessed by either the reward model or the projector. Specifically, we use seeds in $[0, 50)$ to evaluate seen noises and seeds in $[350, 500)$ to evaluate unseen noises.

\noindent \textbf{Compared Baselines.} 
We compare our method against four baselines: the pretrained model, a finetuned model, PAG \cite{PAG}, and AutoGuidance \cite{karras2024guidingdiffusionmodelbad}. The finetuned model is trained on a dataset of images generated with the same noise seeds as our method, with LoRA \cite{hu2021loralowrankadaptationlarge} applied for efficient adaptation. PAG modifies the sampling trajectory by replacing the self-attention maps in the diffusion U-Net with identity matrices. AutoGuidance adjusts classifier-free guidance (CFG) using a weaker model checkpoint to improve image quality. For all training-based baselines, the range of seen noises is kept identical to ours to ensure fairness and avoid introducing extra information.

\begin{table*}[h]
\centering

\resizebox{.98\linewidth}{!}{
\centering

\begin{tabular}{c|c|ccc|ccc}
\toprule
\multirow{2}{*}{Prompt Source}  & \multirow{2}{*}{Method} 
& \multicolumn{3}{c|}{Seen Seeds} & \multicolumn{3}{c}{Unseen Seeds}  \\ \cmidrule{3-8}
& & QwenScore & BERTScore & ImageReward & QwenScore & BERTScore & ImageReward \\
\midrule
\multirow{5}{*}{DrawBench}                 
& Pretrained Model  
& 68.40 \scriptsize$\pm31.01$
& 0.8048 \scriptsize$\pm0.0289$ 
& 1.3199 \scriptsize$\pm0.8725$
& \underline{69.49} \scriptsize$\pm30.33$  
& 0.8038 \scriptsize$\pm0.0295$ 
& 1.2746 \scriptsize$\pm0.8456$
\\ 
& Finetuned Model
& \textbf{70.25} \scriptsize$\pm30.21$  
& \underline{0.8067} \scriptsize$\pm0.0306$  
& \textbf{1.3484} \scriptsize$\pm0.7947$ 
& 69.08 \scriptsize$\pm30.70$ 
& \underline{0.8051} \scriptsize$\pm0.0302$ 
& \underline{1.2811} \scriptsize$\pm0.8436$ 
\\ 
& PAG    
& 50.40 \scriptsize$\pm32.92$  
& 0.7847 \scriptsize$\pm0.0362$  
& 0.1402 \scriptsize$\pm1.4210$  
& 52.24 \scriptsize$\pm32.52$  
& 0.7812 \scriptsize$\pm0.0364$ 
& 0.0678 \scriptsize$\pm1.3965$ 
\\ 
& AutoGuidance  
&  63.07 \scriptsize$\pm33.03$ 
&  0.8017 \scriptsize$\pm0.0369$ 
&  0.8761 \scriptsize$\pm1.2097$ 
&  64.05 \scriptsize$\pm32.63$ 
& 0.7989 \scriptsize$\pm0.0334$ 
& 0.8410 \scriptsize$\pm1.1786$ 
\\ 
& Ours    
&  \underline{70.03} \scriptsize$\pm30.55$ 
&  \textbf{0.8069} \scriptsize$\pm0.0272$ 
&  \underline{1.3289} \scriptsize$\pm0.8042$ 
&  \textbf{70.55} \scriptsize$\pm30.04$ 
&  \textbf{0.8060} \scriptsize$\pm0.0297$ 
&  \textbf{1.3040} \scriptsize$\pm0.8447$ 
\\ \midrule
\multirow{5}{*}{GPT}                 
& Pretrained Model 
& 70.50 \scriptsize$\pm28.85$ 
& \underline{0.8217} \scriptsize$\pm0.0237$ 
& 0.9591 \scriptsize$\pm1.1735$
&  \underline{70.77} \scriptsize$\pm28.96$
& \underline{0.8221} \scriptsize$\pm0.0236$
& 0.9420 \scriptsize$\pm1.1786$ 
\\ 
& Finetuned Model
&  \underline{70.54} \scriptsize$\pm28.97$ 
&  0.8209 \scriptsize$\pm0.0245$ 
&  \textbf{1.0084} \scriptsize$\pm1.1441$
&  70.37 \scriptsize$\pm29.34$
&  0.8219 \scriptsize$\pm0.0235$
&  \underline{0.9779} \scriptsize$\pm1.1599$
\\ 
& PAG    
&  49.91 \scriptsize$\pm32.48$ 
&  0.8048 \scriptsize$\pm0.0300$ 
& -0.1211 \scriptsize$\pm1.3268$
&  48.95 \scriptsize$\pm31.88$
&  0.8024 \scriptsize$\pm0.0292$
&  -0.0719 \scriptsize$\pm1.3346$
\\ 
& AutoGuidance  
&  65.05 \scriptsize$\pm30.36$ 
&  0.8187 \scriptsize$\pm0.0275$ 
&  0.3720 \scriptsize$\pm1.3529$ 
&  65.45 \scriptsize$\pm30.14$
&  0.8173 \scriptsize$\pm0.0270$
& 0.3562 \scriptsize$\pm1.3715$
\\ 
& Ours    
&  \textbf{71.87} \scriptsize$\pm28.24$ 
&  \textbf{0.8226} \scriptsize$\pm0.0224$ 
&  \underline{1.0000} \scriptsize$\pm1.1867$ 
&  \textbf{71.45} \scriptsize$\pm28.45$
& \textbf{0.8228} \scriptsize$\pm0.0234$
& \textbf{1.0017} \scriptsize$\pm1.1580$
\\ \bottomrule
\end{tabular}

}
\caption{
Comparison results of text–image alignment under multiple prompts. Higher values indicate better performance across all evaluation metrics. We report results separately on seen and unseen data, with both mean and standard deviation. The \textbf{bold} and \underline{underline} entries denote the best and second-best results, respectively.
}
\label{table: multiple prompt results}
\end{table*}

\subsection{Single-Prompt Evaluation}\label{subsection: single prompt evaluation}

In this setting, a noise projector is trained to specialize in one specific prompt. Detailed Results are reported in the Appendix. 
For seen data, the finetuned model shows clear improvements, \textit{e.g.}, QwenScore increases by 4.66 on the third prompt. However, such improvements do not extend to unseen data, where the corresponding gain is only 0.18. This is because finetuning alters the parameters of the pretrained model, thereby changing the associated ODE sampler. Without explicit supervision from humans or VLMs, the adapted sampler remains confined to the training distribution and fails to sustain alignment improvements to unseen inputs.
A similar limitation is observed for PAG and AutoGuidance. Both methods locally alter the sampling path—by replacing attention maps or adjusting the negative guidance term—but do so without explicit supervised feedback. Consequently, their performance is unstable and may even fall below that of the pretrained SD model for certain prompts.  
In contrast, our method consistently improves text-image alignment on both seen and unseen data. This advantage arises because the noise projector leverages supervised signals from the VLM to project raw noises into more informative and alignment-friendly ones.

\subsection{Multi-Prompt Evaluation}\label{subsection: multiple prompt evaluation}

The training process of our noise projector naturally supports multiple conditioning prompts as input, allowing a single projector to handle diverse prompts simultaneously. To validate this ability, we select five prompts from each prompt set and mix them to train a noise projector. Evaluation results are reported in Table~\ref{table: multiple prompt results}.  

We could observe that our method consistently improves text–image alignment across mixed prompts. Moreover, in most cases, the noise projector yields smaller standard deviations than baseline methods. This observation is consistent with our key assumption: the projector enhances alignment by mapping raw noise into a distribution enriched with semantic information. Because it is trained on a limited set of noises, the resulting distribution is narrower than the normal Gaussian used in pretrained SD models, leading to outputs that are more similar to each other. Another quantitative evidence supporting this explanation is provided in Section~\ref{subsection: investigation on projected distribution}.

\begin{figure}[h]
    \centering
    \includegraphics[width=\linewidth]{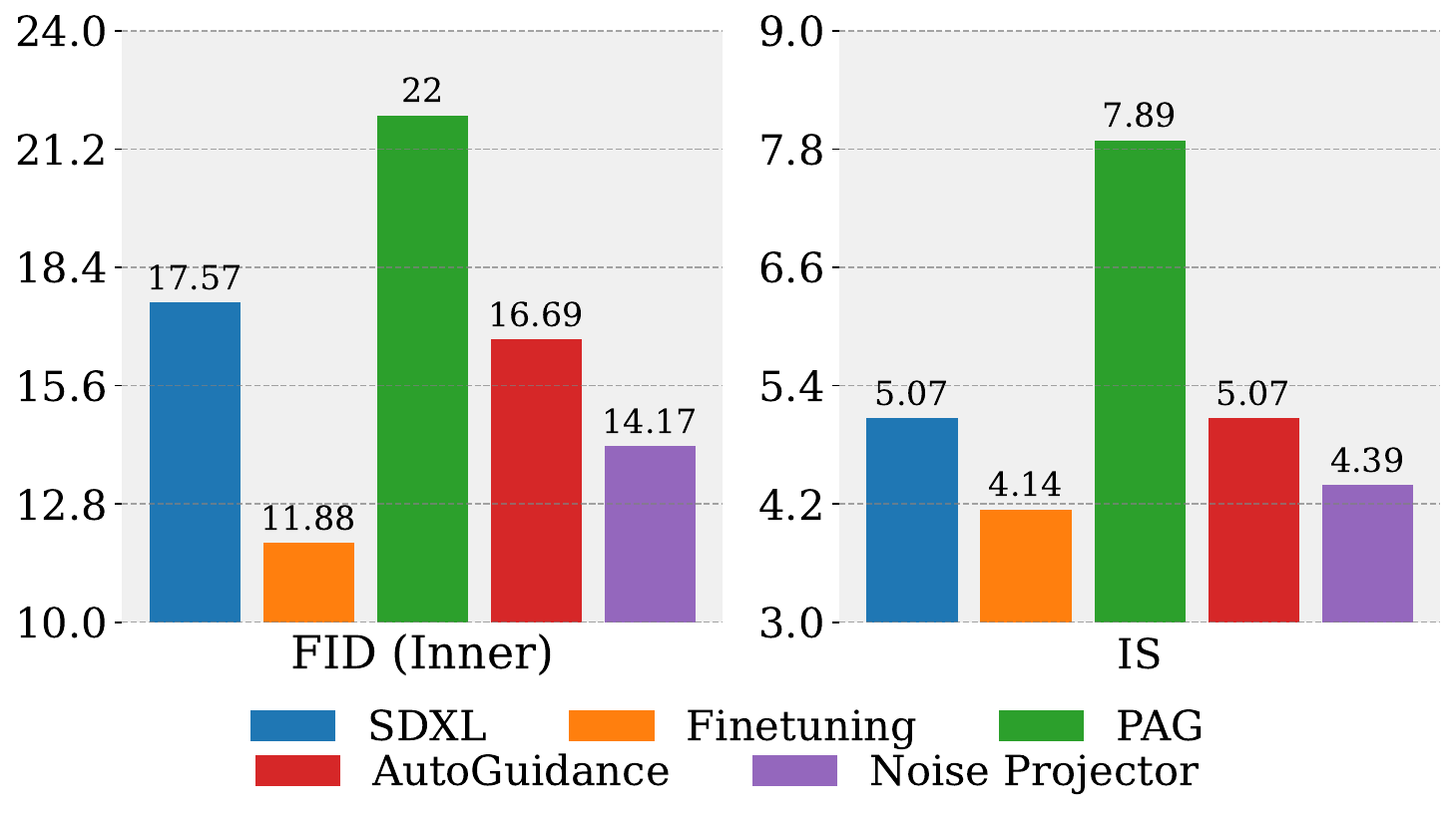}
    \caption{The innerly-computed FID and IS comparison with a single prompt.}
    \label{fig: distribution measure}
\end{figure}

\subsection{Investigation on Projected Distribution}\label{subsection: investigation on projected distribution}

As discussed in Section~\ref{subsection: motivation of projecting noise}, each prompt is paired only with the noise realizations from its training images. The most suitable initial noises for a given prompt may not span the entire Gaussian space. 
Pretrained models tend to favor these noises, yielding well-aligned outputs but failing on less suitable ones. During inference, the ideal noise distribution for a prompt therefore does not necessarily follow the standard Gaussian. 
This analysis motivates our noise projection approach, which we further examine using Fréchet Inception Distance (FID) \cite{FID} and Inception Score (IS) \cite{IS}.
FID assesses similarity and diversity relative to a reference set, and IS measures output diversity from inception features. For a single prompt, 5000 images are generated with unseen noises (seeds 1000–6000). FID is computed by splitting the images into two groups, taking each as reference in turn, and averaging results over 10 runs. IS is computed by dividing the images into 10 splits and measuring the KL divergence between each split and the whole set. In this setting, lower IS and FID indicate reduced diversity, implying a narrower noise distribution. Results are shown in Figure~\ref{fig: distribution measure}.  

There are two observations. First, images produced with our noise projector exhibit lower diversity than those from the pretrained model, confirming that the projector reduces the noise distribution into a narrower form. Second, the finetuned model produces even narrower outputs than our method. This is expected, since training directly on a limited image set pushes generations closer to the training distribution. In contrast, our projector is guided not only by the provided noises but also by signals from the reward model, which emphasize text–image alignment rather than replication. Therefore, its outputs remain more diverse than those of the finetuned model while improving alignment.

\begin{table}[h]
\centering

\resizebox{.98\linewidth}{!}{
\centering

\begin{tabular}{c|ccc}
\toprule
$\tau$ & QwenScore & BERTScore & ImageReward \\
\midrule
100 
& 69.31 \scriptsize$\pm30.26$
& 0.8029 \scriptsize$\pm0.0291$ 
& 1.2669 \scriptsize$\pm0.8465$
\\ 
150
&  70.22 \scriptsize$\pm30.36$
&  0.8046 \scriptsize$\pm0.0298$ 
&  1.2839 \scriptsize$\pm0.8114$
\\ 
200 (default)
&  70.55 \scriptsize$\pm30.04$ 
&  0.8060 \scriptsize$\pm0.0297$ 
&  1.3040 \scriptsize$\pm0.8447$ 
\\ 
250
& 69.98 \scriptsize$\pm30.08$
& 0.8041 \scriptsize$\pm0.0300$ 
& 1.2872 \scriptsize$\pm0.8117$
\\ 
300
& 70.13 \scriptsize$\pm30.04$
& 0.8047 \scriptsize$\pm0.0312$ 
& 1.3017 \scriptsize$\pm0.8503$
\\ \bottomrule
\end{tabular}

}
\caption{Ablation study on $\tau$ with prompts from DrawBench.
}
\label{table: ablation}
\end{table}

\subsection{Sensitivity Analysis}
$\tau$ balances the reward term and the distributional constraint. We study its effect on the noise projector using DrawBench prompts, with results in Table~\ref{table: ablation}. Performance remains stable across a moderate range of $\tau$ (200–300), where the projector consistently outperforms the baseline. However, reducing $\tau$ to a small value (100) leads to clear degradation. This highlights the importance of $\mathcal{L}_{\text{constraint}}$, which regulates the deviation of sampled noises from the standard Gaussian and thereby prevents reward hacking.

\section{Conclusion}
In this paper, we study text–image misalignment in stable diffusion and trace it to a training–inference mismatch: during training, each prompt implicitly induces a prompt-specific noise distribution that allows strong alignment, whereas at inference, initial noise is drawn from a prompt-agnostic Gaussian. Thus, we propose a noise projector that maps random noise to a prompt-aware version. Denoising is then applied on this refined noise for better alignment. 
We train the projector with a quasi-direct preference optimization scheme, and results on both single- and multi-prompt settings show clear gains.
{
    \small
    \bibliographystyle{ieeenat_fullname}
    \bibliography{main}

\begin{thebibliography}{41}
\providecommand{\natexlab}[1]{#1}
\providecommand{\url}[1]{\texttt{#1}}
\expandafter\ifx\csname urlstyle\endcsname\relax
  \providecommand{\doi}[1]{doi: #1}\else
  \providecommand{\doi}{doi: \begingroup \urlstyle{rm}\Url}\fi

\bibitem[Ahn et~al.(2024)Ahn, Cho, Min, Jang, Kim, Kim, Park, Jin, and Kim]{PAG}
Donghoon Ahn, Hyoungwon Cho, Jaewon Min, Wooseok Jang, Jungwoo Kim, SeonHwa Kim, Hyun~Hee Park, Kyong~Hwan Jin, and Seungryong Kim.
\newblock Self-rectifying diffusion sampling with perturbed-attention guidance.
\newblock In \emph{European Conference on Computer Vision}, pages 1--17. Springer, 2024.

\bibitem[Bai et~al.(2025)Bai, Chen, Liu, Wang, Ge, Song, Dang, Wang, Wang, Tang, Zhong, Zhu, Yang, Li, Wan, Wang, Ding, Fu, Xu, Ye, Zhang, Xie, Cheng, Zhang, Yang, Xu, and Lin]{bai2025qwen25vltechnicalreport}
Shuai Bai, Keqin Chen, Xuejing Liu, Jialin Wang, Wenbin Ge, Sibo Song, Kai Dang, Peng Wang, Shijie Wang, Jun Tang, Humen Zhong, Yuanzhi Zhu, Mingkun Yang, Zhaohai Li, Jianqiang Wan, Pengfei Wang, Wei Ding, Zheren Fu, Yiheng Xu, Jiabo Ye, Xi Zhang, Tianbao Xie, Zesen Cheng, Hang Zhang, Zhibo Yang, Haiyang Xu, and Junyang Lin.
\newblock Qwen2.5-vl technical report, 2025.

\bibitem[Black et~al.(2023)Black, Janner, Du, Kostrikov, and Levine]{DDPO}
Kevin Black, Michael Janner, Yilun Du, Ilya Kostrikov, and Sergey Levine.
\newblock Training diffusion models with reinforcement learning.
\newblock \emph{arXiv preprint arXiv:2305.13301}, 2023.

\bibitem[Devlin(2018)]{BERT}
Jacob Devlin.
\newblock Bert: Pre-training of deep bidirectional transformers for language understanding.
\newblock \emph{arXiv preprint arXiv:1810.04805}, 2018.

\bibitem[Eyring et~al.(2024)Eyring, Karthik, Roth, Dosovitskiy, and Akata]{eyring2024reno}
Luca Eyring, Shyamgopal Karthik, Karsten Roth, Alexey Dosovitskiy, and Zeynep Akata.
\newblock Reno: Enhancing one-step text-to-image models through reward-based noise optimization.
\newblock \emph{Neural Information Processing Systems (NeurIPS)}, 2024.

\bibitem[Guo et~al.(2024)Guo, Liu, Cui, Li, Yang, and Huang]{guo2024initno}
Xiefan Guo, Jinlin Liu, Miaomiao Cui, Jiankai Li, Hongyu Yang, and Di Huang.
\newblock Initno: Boosting text-to-image diffusion models via initial noise optimization.
\newblock In \emph{Proceedings of the IEEE/CVF Conference on Computer Vision and Pattern Recognition}, pages 9380--9389, 2024.

\bibitem[He et~al.(2021)He, Liu, Gao, and Chen]{he2021deberta}
Pengcheng He, Xiaodong Liu, Jianfeng Gao, and Weizhu Chen.
\newblock Deberta: Decoding-enhanced bert with disentangled attention.
\newblock In \emph{International Conference on Learning Representations}, 2021.

\bibitem[Heusel et~al.(2018)Heusel, Ramsauer, Unterthiner, Nessler, and Hochreiter]{FID}
Martin Heusel, Hubert Ramsauer, Thomas Unterthiner, Bernhard Nessler, and Sepp Hochreiter.
\newblock Gans trained by a two time-scale update rule converge to a local nash equilibrium, 2018.

\bibitem[Ho and Salimans(2022)]{CFG}
Jonathan Ho and Tim Salimans.
\newblock Classifier-free diffusion guidance, 2022.

\bibitem[Ho et~al.(2020)Ho, Jain, and Abbeel]{ddpm}
Jonathan Ho, Ajay Jain, and Pieter Abbeel.
\newblock Denoising diffusion probabilistic models.
\newblock \emph{Advances in Neural Information Processing Systems}, 33:\penalty0 6840--6851, 2020.

\bibitem[Hu et~al.(2021)Hu, Shen, Wallis, Allen-Zhu, Li, Wang, Wang, and Chen]{hu2021loralowrankadaptationlarge}
Edward~J. Hu, Yelong Shen, Phillip Wallis, Zeyuan Allen-Zhu, Yuanzhi Li, Shean Wang, Lu Wang, and Weizhu Chen.
\newblock Lora: Low-rank adaptation of large language models, 2021.

\bibitem[Hu et~al.(2025{\natexlab{a}})Hu, Zhang, Chen, Kuang, Li, Gao, Xiao, Wang, and Zhu]{hu2025towards}
Zijing Hu, Fengda Zhang, Long Chen, Kun Kuang, Jiahui Li, Kaifeng Gao, Jun Xiao, Xin Wang, and Wenwu Zhu.
\newblock Towards better alignment: Training diffusion models with reinforcement learning against sparse rewards.
\newblock \emph{arXiv preprint arXiv:2503.11240}, 2025{\natexlab{a}}.

\bibitem[Hu et~al.(2025{\natexlab{b}})Hu, Zhang, and Kuang]{hu2025d}
Zijing Hu, Fengda Zhang, and Kun Kuang.
\newblock D-fusion: Direct preference optimization for aligning diffusion models with visually consistent samples.
\newblock \emph{arXiv preprint arXiv:2505.22002}, 2025{\natexlab{b}}.

\bibitem[Karras et~al.(2022)Karras, Aittala, Aila, and Laine]{EDM}
Tero Karras, Miika Aittala, Timo Aila, and Samuli Laine.
\newblock Elucidating the design space of diffusion-based generative models.
\newblock \emph{Advances in neural information processing systems}, 35:\penalty0 26565--26577, 2022.

\bibitem[Karras et~al.(2024)Karras, Aittala, Kynkäänniemi, Lehtinen, Aila, and Laine]{karras2024guidingdiffusionmodelbad}
Tero Karras, Miika Aittala, Tuomas Kynkäänniemi, Jaakko Lehtinen, Timo Aila, and Samuli Laine.
\newblock Guiding a diffusion model with a bad version of itself, 2024.

\bibitem[Kynk{\"a}{\"a}nniemi et~al.(2024)Kynk{\"a}{\"a}nniemi, Aittala, Karras, Laine, Aila, and Lehtinen]{kynkaanniemi2024applying}
Tuomas Kynk{\"a}{\"a}nniemi, Miika Aittala, Tero Karras, Samuli Laine, Timo Aila, and Jaakko Lehtinen.
\newblock Applying guidance in a limited interval improves sample and distribution quality in diffusion models.
\newblock \emph{Advances in Neural Information Processing Systems}, 37:\penalty0 122458--122483, 2024.

\bibitem[Labs(2024)]{flux2024}
Black~Forest Labs.
\newblock Flux.
\newblock \url{https://github.com/black-forest-labs/flux}, 2024.

\bibitem[Lu et~al.(2022)Lu, Zhou, Bao, Chen, Li, and Zhu]{lu2022dpm}
Cheng Lu, Yuhao Zhou, Fan Bao, Jianfei Chen, Chongxuan Li, and Jun Zhu.
\newblock Dpm-solver: A fast ode solver for diffusion probabilistic model sampling in around 10 steps.
\newblock \emph{arXiv preprint arXiv:2206.00927}, 2022.

\bibitem[Ma et~al.(2025)Ma, Tong, Jia, Hu, Su, Zhang, Yang, Li, Jaakkola, Jia, et~al.]{ma2025inference}
Nanye Ma, Shangyuan Tong, Haolin Jia, Hexiang Hu, Yu-Chuan Su, Mingda Zhang, Xuan Yang, Yandong Li, Tommi Jaakkola, Xuhui Jia, et~al.
\newblock Inference-time scaling for diffusion models beyond scaling denoising steps.
\newblock \emph{arXiv preprint arXiv:2501.09732}, 2025.

\bibitem[Miao et~al.(2025)Miao, Li, Wang, Zhang, Sun, Wang, and Zhu]{miao2025noise}
Boming Miao, Chunxiao Li, Xiaoxiao Wang, Andi Zhang, Rui Sun, Zizhe Wang, and Yao Zhu.
\newblock Noise diffusion for enhancing semantic faithfulness in text-to-image synthesis.
\newblock In \emph{Proceedings of the Computer Vision and Pattern Recognition Conference}, pages 23575--23584, 2025.

\bibitem[OpenAI(2024)]{openai2024gpt4ocard}
OpenAI.
\newblock Gpt-4o system card, 2024.

\bibitem[Podell et~al.(2024)Podell, English, Lacey, Blattmann, Dockhorn, M{\"u}ller, Penna, and Rombach]{SDXL}
Dustin Podell, Zion English, Kyle Lacey, Andreas Blattmann, Tim Dockhorn, Jonas M{\"u}ller, Joe Penna, and Robin Rombach.
\newblock {SDXL}: Improving latent diffusion models for high-resolution image synthesis.
\newblock In \emph{The Twelfth International Conference on Learning Representations}, 2024.

\bibitem[Rombach et~al.(2022)Rombach, Blattmann, Lorenz, Esser, and Ommer]{stable-diffusion}
Robin Rombach, Andreas Blattmann, Dominik Lorenz, Patrick Esser, and Bj{\"o}rn Ommer.
\newblock High-resolution image synthesis with latent diffusion models.
\newblock In \emph{Proceedings of the IEEE/CVF conference on computer vision and pattern recognition}, pages 10684--10695, 2022.

\bibitem[Saharia et~al.(2022)Saharia, Chan, Saxena, Li, Whang, Denton, Ghasemipour, Gontijo~Lopes, Karagol~Ayan, Salimans, et~al.]{drawbench}
Chitwan Saharia, William Chan, Saurabh Saxena, Lala Li, Jay Whang, Emily~L Denton, Kamyar Ghasemipour, Raphael Gontijo~Lopes, Burcu Karagol~Ayan, Tim Salimans, et~al.
\newblock Photorealistic text-to-image diffusion models with deep language understanding.
\newblock \emph{Advances in Neural Information Processing Systems}, 35:\penalty0 36479--36494, 2022.

\bibitem[Salimans et~al.(2016)Salimans, Goodfellow, Zaremba, Cheung, Radford, and Chen]{IS}
Tim Salimans, Ian Goodfellow, Wojciech Zaremba, Vicki Cheung, Alec Radford, and Xi Chen.
\newblock Improved techniques for training gans, 2016.

\bibitem[Song et~al.(2020)Song, Meng, and Ermon]{DDIM}
Jiaming Song, Chenlin Meng, and Stefano Ermon.
\newblock Denoising diffusion implicit models.
\newblock \emph{arXiv preprint arXiv:2010.02502}, 2020.

\bibitem[Song et~al.(2021)Song, Sohl-Dickstein, Kingma, Kumar, Ermon, and Poole]{score-based-method}
Yang Song, Jascha Sohl-Dickstein, Diederik~P Kingma, Abhishek Kumar, Stefano Ermon, and Ben Poole.
\newblock Score-based generative modeling through stochastic differential equations.
\newblock In \emph{International Conference on Learning Representations}, 2021.

\bibitem[Sundaram et~al.(2024)Sundaram, Pal, Chauhan, Agarwal, and Karanam]{sundaram2024coconoattentioncontrastandcompleteinitial}
Aravindan Sundaram, Ujjayan Pal, Abhimanyu Chauhan, Aishwarya Agarwal, and Srikrishna Karanam.
\newblock Cocono: Attention contrast-and-complete for initial noise optimization in text-to-image synthesis, 2024.

\bibitem[Tong et~al.(2025)Tong, Zhang, Tang, Gao, Huang, Lyu, Xiao, and Kuang]{tong2025latent}
Yunze Tong, Fengda Zhang, Zihao Tang, Kaifeng Gao, Kai Huang, Pengfei Lyu, Jun Xiao, and Kun Kuang.
\newblock Latent score-based reweighting for robust classification on imbalanced tabular data.
\newblock In \emph{Forty-second International Conference on Machine Learning}, 2025.

\bibitem[Xiao et~al.(2024)Xiao, Yin, Freeman, Durand, and Han]{fastcomposer}
Guangxuan Xiao, Tianwei Yin, William~T Freeman, Fr{\'e}do Durand, and Song Han.
\newblock Fastcomposer: Tuning-free multi-subject image generation with localized attention.
\newblock \emph{International Journal of Computer Vision}, pages 1--20, 2024.

\bibitem[Xu et~al.(2023)Xu, Liu, Wu, Tong, Li, Ding, Tang, and Dong]{imagereward}
Jiazheng Xu, Xiao Liu, Yuchen Wu, Yuxuan Tong, Qinkai Li, Ming Ding, Jie Tang, and Yuxiao Dong.
\newblock Imagereward: Learning and evaluating human preferences for text-to-image generation.
\newblock \emph{Advances in Neural Information Processing Systems}, 36:\penalty0 15903--15935, 2023.

\bibitem[Yang et~al.(2024{\natexlab{a}})Yang, Tang, Zhu, Chen, Shen, and Wu]{yang2024mitigating}
Jinluan Yang, Anke Tang, Didi Zhu, Zhengyu Chen, Li Shen, and Fei Wu.
\newblock Mitigating the backdoor effect for multi-task model merging via safety-aware subspace.
\newblock \emph{arXiv preprint arXiv:2410.13910}, 2024{\natexlab{a}}.

\bibitem[Yang et~al.(2024{\natexlab{b}})Yang, Zhang, Chen, Xiao, Wang, Wu, and Kuang]{yang2024discovering}
Jinluan Yang, Ruihao Zhang, Zhengyu Chen, Teng Xiao, Yueyang Wang, Fei Wu, and Kun Kuang.
\newblock Discovering invariant neighborhood patterns for heterophilic graphs.
\newblock \emph{arXiv preprint arXiv:2403.10572}, 2024{\natexlab{b}}.

\bibitem[Yang et~al.(2025{\natexlab{a}})Yang, Chen, Xiao, Lin, Zhang, and Kuang]{yang2025leveraging}
Jinluan Yang, Zhengyu Chen, Teng Xiao, Yong Lin, Wenqiao Zhang, and Kun Kuang.
\newblock Leveraging invariant principle for heterophilic graph structure distribution shifts.
\newblock In \emph{Proceedings of the ACM on Web Conference 2025}, pages 1196--1204, 2025{\natexlab{a}}.

\bibitem[Yang et~al.(2025{\natexlab{b}})Yang, Jin, Tang, Shen, Zhu, Chen, Zhao, Wang, Cui, Zhang, et~al.]{yang2025mix}
Jinluan Yang, Dingnan Jin, Anke Tang, Li Shen, Didi Zhu, Zhengyu Chen, Ziyu Zhao, Daixin Wang, Qing Cui, Zhiqiang Zhang, et~al.
\newblock Mix data or merge models? balancing the helpfulness, honesty, and harmlessness of large language model via model merging.
\newblock \emph{arXiv preprint arXiv:2502.06876}, 2025{\natexlab{b}}.

\bibitem[Yang et~al.(2025{\natexlab{c}})Yang, Zhang, Chen, Wu, and Kuang]{yang2025unifying}
Jinluan Yang, Ruihao Zhang, Zhengyu Chen, Fei Wu, and Kun Kuang.
\newblock Unifying adversarial perturbation for graph neural networks.
\newblock \emph{arXiv preprint arXiv:2509.00387}, 2025{\natexlab{c}}.

\bibitem[Ye et~al.(2023)Ye, Zhang, Liu, Han, and Yang]{ip-adapter}
Hu Ye, Jun Zhang, Sibo Liu, Xiao Han, and Wei Yang.
\newblock Ip-adapter: Text compatible image prompt adapter for text-to-image diffusion models.
\newblock 2023.

\bibitem[Zhang et~al.(2024)Zhang, Zhang, Srinivasan, Shen, Qin, Faloutsos, Rangwala, and Karypis]{tabsyn}
Hengrui Zhang, Jiani Zhang, Balasubramaniam Srinivasan, Zhengyuan Shen, Xiao Qin, Christos Faloutsos, Huzefa Rangwala, and George Karypis.
\newblock Mixed-type tabular data synthesis with score-based diffusion in latent space.
\newblock In \emph{The twelfth International Conference on Learning Representations}, 2024.

\bibitem[Zhang et~al.(2023)Zhang, Rao, and Agrawala]{controlnet}
Lvmin Zhang, Anyi Rao, and Maneesh Agrawala.
\newblock Adding conditional control to text-to-image diffusion models, 2023.

\bibitem[Zhou et~al.(2025)Zhou, Shao, Bai, Zhang, Xu, Han, and Xie]{zhou2025golden}
Zikai Zhou, Shitong Shao, Lichen Bai, Shufei Zhang, Zhiqiang Xu, Bo Han, and Zeke Xie.
\newblock Golden noise for diffusion models: A learning framework.
\newblock In \emph{International Conference on Computer Vision}, 2025.

\bibitem[Zhu et~al.(2025)Zhu, Song, Shen, Zhao, Yang, Zhang, and Wu]{zhu2025remedy}
Didi Zhu, Yibing Song, Tao Shen, Ziyu Zhao, Jinluan Yang, Min Zhang, and Chao Wu.
\newblock Remedy: Recipe merging dynamics in large vision-language models.
\newblock In \emph{The Thirteenth International Conference on Learning Representations}, 2025.

\end{thebibliography}
}


\end{document}